\documentclass[10pt,twocolumn,letterpaper]{article}

\usepackage[pagenumbers]{cvpr} %

\usepackage[dvipsnames]{xcolor}

\definecolor{cvprblue}{rgb}{0.21,0.49,0.74}
\usepackage[pagebackref,breaklinks,colorlinks,citecolor=cvprblue]{hyperref}
\usepackage[accsupp]{axessibility}
\def\paperID{545} %
\def\confName{CVPR}
\def\confYear{2024}

\usepackage{amsfonts}
\usepackage{bm}
\usepackage{amsmath}
\usepackage{mathtools}
\usepackage{multirow}
\usepackage{booktabs}
\usepackage{caption}
\usepackage{enumitem}
\usepackage{wrapfig,lipsum,booktabs}
\usepackage{caption}
\usepackage{bbm}
\usepackage{multirow}
\usepackage{pifont}
\usepackage[linesnumbered,ruled,vlined]{algorithm2e}

\SetCommentSty{mycommfont}
\SetAlCapFnt{\small} 
\usepackage{etoc}
\SetAlgorithmName{Algo}{Algo}{}

\renewcommand{\paragraph}[1]{{\vspace{1mm}\noindent \bf #1}.}

\newcommand{\reals}{\mathbb{R}}

\newcommand{\imaget}{\bs{I}_{t}}
\newcommand{\images}{\bs{I}_{1:T}}

\newcommand{\imagetn}{\bs{I}_{t + 1}}

\newcommand{\name}{\texttt{SimXR}}
\newcommand{\policyphc}{{\pi_{\text{PHC}}}}

\newcommand{\policyours}{{\pi_{\text{\name}}}}

\newcommand{\rewardfunc}{\bs{\mathcal{R}}}

\newcommand{\refps}{{\bs{\hat{q}}_{1:T}}}

\newcommand{\refpzero}{{\bs{\hat{q}}_{0}}}

\newcommand{\refv}{{\bs{\hat{\dot{q}}}_{t}}}

\newcommand{\refpn}{{\bs{\hat{q}}_{t+1}}}

\newcommand{\reftn}{{\bs{\hat{p}}_{t+1}}}

\newcommand{\refrn}{{\bs{\hat{\theta}}_{t+1}}}

\newcommand{\kinp}{{\bs{\tilde{q}}_{t}}}

\newcommand{\kinpn}{{\bs{\Tilde{q}}_{t+1}}}

\newcommand{\simp}{{\bs{{q}}_{t}}}

\newcommand{\simv}{{\bs{\dot{q}}_{t}}}
\newcommand{\simvs}{{\bs{\dot{q}}_{1:T}}}
\newcommand{\simav}{{\bs{{\omega}}_{t}}}
\newcommand{\simlv}{{\bs{v}_{t}}}

\newcommand{\simr}{{\bs{{\theta}}_{t}}}
\newcommand{\simt}{{\bs{{p}}_{t}}}
\newcommand{\goalstate}{{\bs{s}^{\text{g}}_t}}

\newcommand{\goalstateimitate}{{\bs{s}^{\text{g-mimic}}_t}}

\newcommand{\selfstate}{{\bs{s}^{\text{p}}_t}}
\newcommand{\selfstates}{{\bs{s}^{\text{p}}_{1:T}}}
\newcommand{\state}{{\bs{s}_t}}
\newcommand{\action}{{\bs{a}_t}}
\newcommand{\phcplusaction}{{\bs{a}^{\text{PHC+}}_t}}
\newcommand{\actiongt}{{\hat{\bs{a}}_t}}
\newcommand{\actionphc}{{\bs{a}^{\text{PHC}}_t}}

\newcommand{\motiondata}{\bs{\hat{Q}}}

\newcommand{\hardmotiondata}{\bs{\hat{Q}}_{\text{hard}}}

\newcommand{\mpjpe}{E_\text{mpjpe}}

\newcommand{\gmpjpe}{E_\text{g-mpjpe}}
\newcommand{\pampjpe}{E_\text{pa-mpjpe}}
\newcommand{\acc}{E_\text{acc}}
\newcommand{\vel}{E_\text{vel}}
\newcommand{\success}{\text{Succ}}

\newcommand{\trajs}{\bs{q}^{\tau}_{1:T}}
\newcommand{\trajn}{\bs{q}^{\tau}_{t+1}}
\newcommand{\traj}{\simp^{\tau}}

\newcommand{\trajtn}{\bs{p}^{\tau}_{t + 1}}
\newcommand{\trajrn}{\bs{\theta}^{\tau}_{t+1}}

\newcommand{\resnet}{\mathcal{F}}
\newcommand{\imagefeat}{\bs{\phi}_t}
\newcommand{\trajfeat}{\bs{\psi}_t}

\newcommand{\bs}[1]{\boldsymbol{#1}}
\newcommand{\cmark}{\ding{51}}%
\newcommand{\xmark}{\ding{55}}%

\newcommand\blfootnote[1]{%
  \begingroup
  \renewcommand\thefootnote{}\footnote{#1}%
  \addtocounter{footnote}{-1}%
  \endgroup
}

\title{Real-Time Simulated Avatar from Head-Mounted Sensors}

\author{
Zhengyi Luo$^{1,2}$
\quad
Jinkun Cao$^{2}$
\quad
Rawal Khirodkar$^{1}$
\quad
Alexander Winkler$^{1}$ 
\quad
Jing Huang$^{1}$ \\
\quad
Kris Kitani$^{1,2,*}$ 
\quad
Weipeng Xu$^{1,*}$  \\
$^{1}$Reality Labs Research, Meta; $^{2}$Carnegie Mellon University\\
{\tt\small \href{https://zhengyiluo.github.io/SimXR/}{https://zhengyiluo.github.io/SimXR/}  }}

\begin{document}

\twocolumn[{
\renewcommand\twocolumn[1][]{#1}%
\maketitle
\vspace{-0.4in}
\begin{center}
    \centering
    \includegraphics[width=1\textwidth]{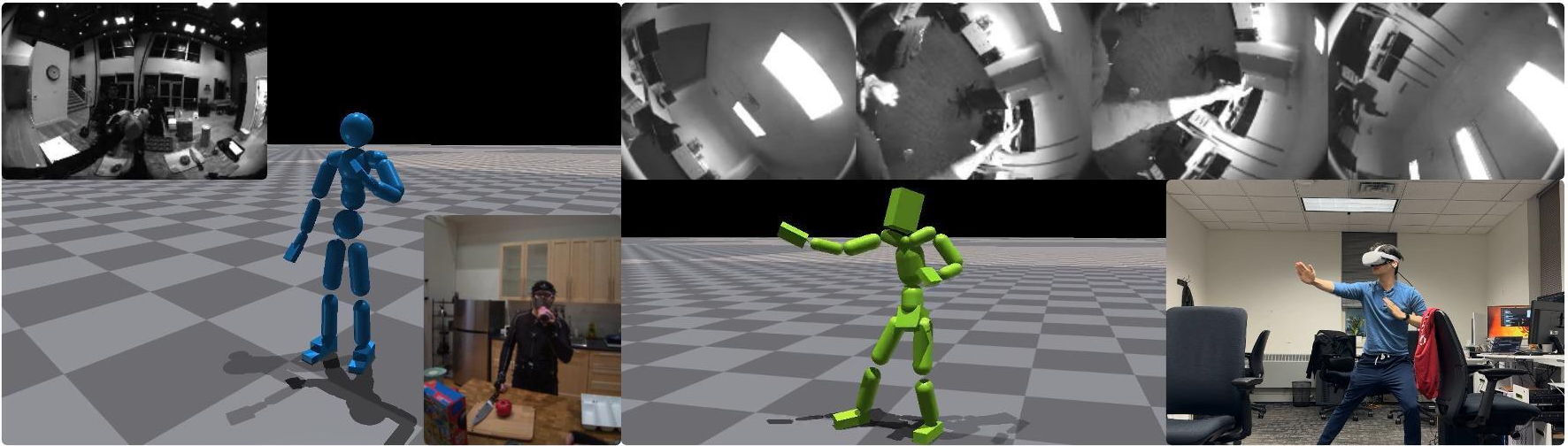}
    \captionof{figure}{\footnotesize{Avatar control using \textit{$\name$} on real world AR/VR headsets. (\textit{Left}): An indoor kitchen setting using AR headset. $\name$ controls the humanoid using headset pose and visual input from two front-facing cameras. (\textit{Right}): An office setting using VR headset (Quest 2). Humanoid motion is driven by the headset pose, two side-facing and two up-facing cameras.}}
    \label{fig:teaser}
\end{center}%
}]
\maketitle
\begin{abstract}
\vspace{-5mm}
    We present \textbf{$\name$}, a method for controlling a simulated avatar from information (headset pose and cameras) obtained from AR / VR headsets. Due to the challenging viewpoint of head-mounted cameras, the human body is often clipped out of view, making traditional image-based egocentric pose estimation challenging. On the other hand, headset poses provide valuable information about overall body motion, but lack fine-grained details about the hands and feet. To synergize headset poses with cameras, we control a humanoid to track headset movement while analyzing input images to decide body movement. When body parts are seen, the movements of hands and feet will be guided by the images; when unseen, the laws of physics guide the controller to generate plausible motion. We design an end-to-end method that does not rely on any intermediate representations and learns to directly map from images and headset poses to humanoid control signals. To train our method, we also propose a large-scale synthetic dataset created using camera configurations compatible with a commercially available VR headset (Quest 2) and show promising results on real-world captures. To demonstrate the applicability of our framework, we also test it on an AR headset with a forward-facing camera. 

\end{abstract}

\etocdepthtag.toc{mtchapter}
\etocsettagdepth{mtchapter}{subsection}
\etocsettagdepth{mtappendix}{none}

\section{Introduction}
\blfootnote{\scriptsize{$^*$Equal advising.}}    
From the sensor streams captured by a head-mounted device (AR / VR, or XR headsets), we aim to control a simulated humanoid / avatar to follow the wearer's global 3D body pose in real-time, as shown in Fig.\ref{fig:teaser}. This could be applied to animating virtual avatars in mixed reality, games, and potentially teleoperating humanoid robots \cite{he2024learning}. However, the sensor suite of commercially available head-mounted devices is rarely designed for full-body pose estimation. Their cameras are often facing forward (\eg Aria glasses \cite{Somasundaram2023ProjectAA}) or on the side (\eg Meta Quest \cite{UnknownUnknown-vg}) and are used mainly for Simultaneous Location and Mapping (SLAM) and hand tracking. Thus, the body is seen from extreme and distorted viewpoints from these cameras.

These challenges have led to research on vision-based egocentric pose estimation to create scenarios with more favorable camera placement (\eg fisheye cameras directly pointing downward \cite{Wang2021-qg, Xu2018-qj, Akada2022-zq, ZhaoUnknown-nl, Wang2022-ph, Tome2020-od}), where more body parts can be observed. These camera views are often unrealistic and hard to recreate in the real-world: a camera protruding out and facing downward could be out of the budget or break the aesthetics of the product.  As there is no standard for these heterogeneous camera specifications, it is difficult to collect large-scale data. Using synthetic data \cite{Tome2020-od, Akada2022-zq} could alleviate this problem to some extent, but real-world recreation of the camera specification used in rendering is still challenging, exacerbating the sim-to-real gap. 

Another line of work instead uses head tracking to infer full-body motion. The $6$ degree-of-freedom (DoF) headset pose, being considerably lower in dimensionality compared to images, is easily accessible from extended reality (XR) headsets. However, the headset pose alone contains insufficient information on full body movement, so previous work either formulates the task as a generative one \cite{Li2022-ud} or relies on action labels \cite{Luo2021-gu} to further constrain the solution space. Adding VR controllers as input can provide additional information on hand movement and can lead to a more stable estimate of full body pose, but the VR controller is not always available \cite{ponton2022combining, UnknownUnknown-ja, Aliakbarian2023-tc}, especially for light-weight AR glasses with forward-facing cameras. These methods are also primarily kinematics-based \cite{Jiang2022-cb, Aliakbarian2023-tc, Aliakbarian2022-mk, Du_undated-ab}, focusing on motion estimation without taking into account the underlying forces. As a result, they cause floating and penetration problems, especially because feet are often unobserved. 
 
To combat these issues, physics simulation \cite{Winkler2022-bv} and environmental cues \cite{Lee2023-tw} have been used to create plausible foot motion. Leveraging the laws of physics can significantly improve motion realism and force the simulated character to adopt a viable foot movement. However, incorporating physics introduces the additional challenge of humanoid control--humanoids need to be balanced and track user movement at all times. Most physics-based methods are learned using reinforcement learning (RL) and require millions (sometimes billions) of environment interactions. If each interaction requires processing of the raw image input, the computational cost would be significant. As a result, approaches that use vision and simulated avatar often use a low-dimensional intermediate representation, such as pre-computed image features \cite{Yuan2019-qp}  or kinematic poses \cite{Luo2022-ux, Luo2021-gu, Yuan2021-rl}. Although a controller can be trained to consume these intermediate features, this approach creates a disconnect between the visual and control components, where the visual component does not receive adequate feedback during training from physics simulation. 

\begin{figure}
\vspace{-1.5mm}
\begin{center}
\includegraphics[width=0.5\textwidth]{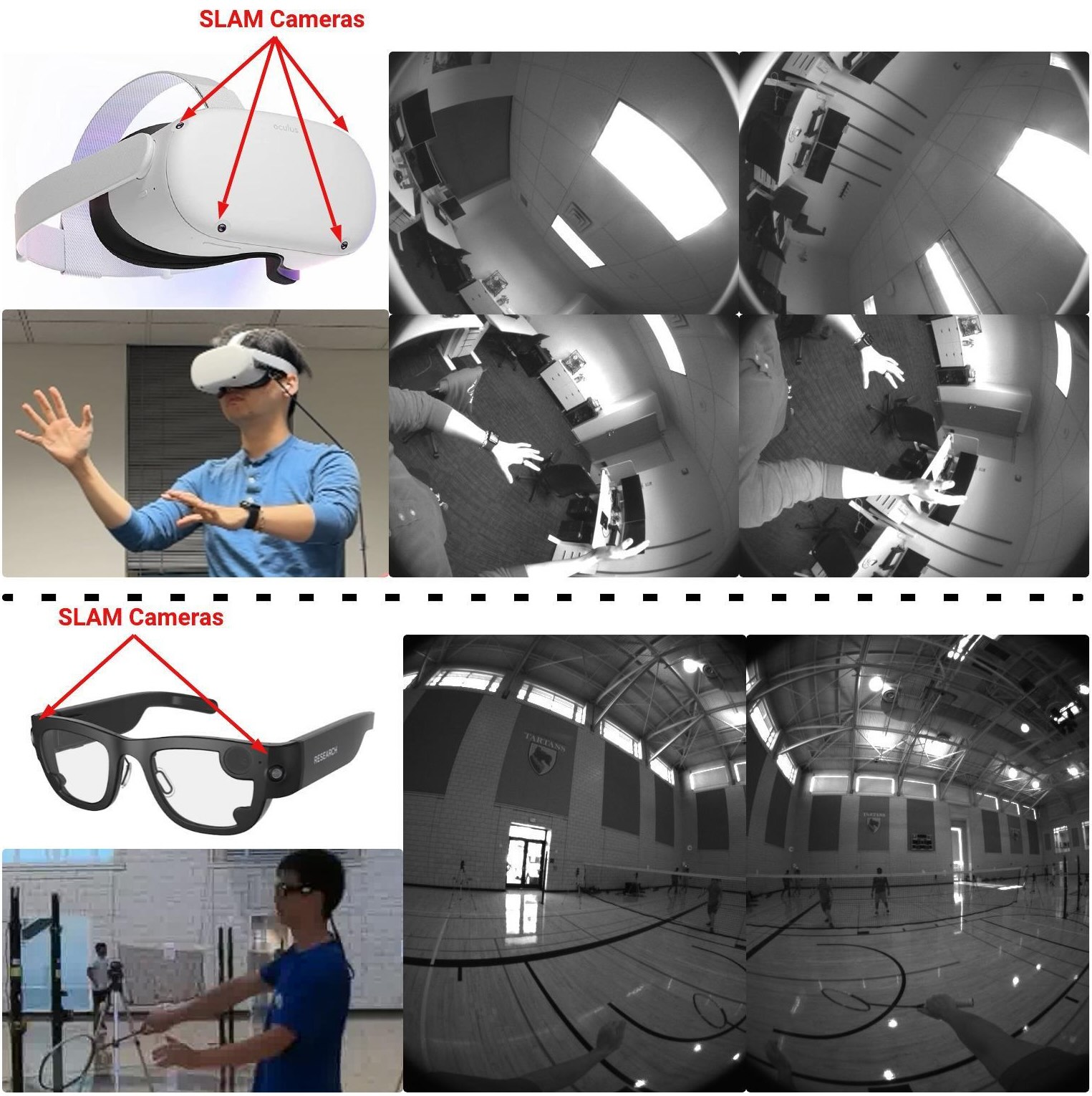}
\end{center}
\vspace{-5mm}
   \caption{\small{SimXR framework applied to two AR/VR devices. (\textit{Top}): Quest 2~\cite{UnknownUnknown-cx} headset with 4 SLAM cameras, two facing upward and two downward. (\textit{Bottom}): Aria glass~\cite{UnknownUnknown-br} with two forward-facing SLAM cameras. Both devices provides 6DoF headset tracking in real-time. }}
\vspace{-3.5mm}
\label{fig:hardware}
\end{figure}

In this work, we demonstrate the feasibility of an end-to-end simulated avatar control framework for XR headsets. Our approach, \textit{\textbf{SIM}ulated Avatar from \textbf{XR} sensors} ($\name$), directly maps the input signals to joint actuation without relying on intermediate representations such as body pose or 2D keypoints.
Our key design choice is to distill from a pre-trained motion imitator to learn the mapping from input to control signals, which enables efficient learning from vision input. Due to its simplistic design and learning framework, our approach is compatible with a diverse selection of smart headsets, ranging from VR goggles to lightweight AR headsets (as shown in Fig.\ref{fig:hardware}). Since no dataset exists for the camera configurations of commercially available VR headsets, we also propose a large-scale synthetic dataset ($2216$k frames) and a real-world dataset ($40$k frames) for testing and show that our method can be applied to real-world and real-time use cases.

To summarize, our contributions are: (1) we design a method to use simulated humanoids to estimate global full-body pose using images and headset pose from an XR headset (front-facing AR cameras or side-facing VR ones); (2) we demonstrate the feasibility of learning an end-to-end controller to directly map from input sensor features to control signals through distillation; (3) we contribute large-scale synthetic and real-world datasets with commercially available VR headset configuration for future research.

\section{Related Work}

\paragraph{Pose Estimation from Head-Mounted Sensors}
Due to the lack of dataset using commercially available devices, egocentric pose estimation has been studied using synthetic data \cite{Akada2022-zq, Tome2020-od}, and small-scale real-world data captured from custom camera rigs \cite{rhodin2016egocap, jiang2017seeing, Xu2018-qj, Yuan2018-ft, Wang2021-qg}. Among them, EgoCap\cite{rhodin2016egocap} uses two downward facing cameras protruding from the helmet, while Jiang \etal \cite{jiang2017seeing} uses a chest-mounted camera. Later, Mo2Cap2 \cite{Xu2018-qj} and SelfPose \cite{Tome2020-od} use a head-mounted downward-facing fisheye camera for pose estimation in the camera's coordinate system. All use synthetic data for training, while Mo2Cap2 also captures a real-world dataset (10k frames) for testing. Wang \etal \cite{Wang2021-qg, Wang2022-ph} extends a similar setup for global 3D body pose estimation by extracting head movement from video and optimization-based global pose refinement. To combat the lack of large-scale dataset, UnrealEgo \cite{Akada2022-zq} uses a dual fish-eye camera setup to generate synthetic data using a game engine. All of the above settings have cameras that directly point downward at the human body. However, cameras on commercially available XR headsets often do not have a dedicated body camera and only have monochrome SLAM or hand-tracking cameras. For VR devices, these cameras point to the side and have very limited visibility of the hands and feet. For AR glasses, cameras often point forward and provide only fleeting hand visibility. Due to challenging viewpoints, some work uses head tracking as an alternative~\cite{Luo2021-gu, Li2022-ud, Yuan2018-ft, Yuan2019-qp, ponton2022combining, Winkler2022-bv, Winkler2022-bv} for pose estimation. Among them, Egopose~\cite{Yuan2019-qp} estimates locomotion, while KinPoly~\cite{Luo2021-gu} extends it to action-conditioned pose estimation. EgoEgo \cite{Li2022-ud} proposes the first general-purpose pose estimator that uses only the head pose. While they utilize front-facing images, the images are used to extract the head pose, rather than to provide information about the body pose. 
In this work, we take advantage of both camera views and headset pose for full-body avatar control.

\paragraph{Simulated Humanoid Motion Imitation} Motion imitation is an important humanoid control task that has seen steady progress in recent years \cite{Al_Borno_undated-uv, Peng2018-fu, Peng2019-kf, Chentanez2018-cw, Won2020-lb, Yuan2020-fp, Fussell2021-jh, Luo2023-ft, Wang2020-qi}. Since no ground-truth data exist of human joint actuation and physics simulators are often nondifferentiable, a policy / imitator / controller is often trained to track / imitate / mimic human motion using deep reinforcement learning (RL). Although methods such as SuperTrack \cite{Fussell2021-jh} and DiffMimic \cite{Ren2023-ho} explore more efficient ways than RL to train imitators, learning a robust policy to track a large amount of human motion remains challenging. Nonetheless, from policies that can track a single clip of motion \cite{Peng2018-fu}, to large-scale datasets \cite{Won2020-lb, Luo2023-ft}, the applicability of motion imitators to downstream tasks grows. Previously, UHC \cite{Luo2021-gu}, a motion imitator based on an external non-physical force \cite{Yuan2020-fp}, has been used for egocentric \cite{Luo2021-gu} and third-person scene-aware \cite{Luo2022-ux} pose estimation. Its follow-up, PHC \cite{Luo2023-ft}, removes the dependency on the non-physical force.

\paragraph{Simulated Humanoid Control for Pose Estimation}
Our work follows recent advances \cite{Yuan2021-rl, Luo2022-ux, Huang2022-bc, Gong2022-sv, Gartner2022-px, WangUnknown-mx} in using the laws of physics as a strong prior to estimating full-body motion. Mapping directly from images to humanoid control signals is quite challenging due to the complex dynamics of a humanoid and the diversity in natural images. 
Training motion controllers using RL is also notoriously sample inefficient, forcing some vision-based robotic approaches to use distillation \cite{Zhuang2023-nf} or very small images \cite{Merel2020-qm}. Therefore, most physics-based methods separate the problem into two distinct components: image-based pose estimation and humanoid motion imitation. First, the pose is estimated from the images using an off-the-shelf body pose \cite{Kocabas2020-vq} or a keypoint detector \cite{Geng2021-xr}. The estimated pose is then fed to a pre-trained imitator for further refinement \cite{Yuan2021-rl, Luo2022-ux}, sampling-based control \cite{Huang2022-bc}, or co-training \cite{Gong2022-sv}. Some methods also employ trajectory optimization \cite{Rempe2020-wl, Shimada2020-wv, Xie2021-ic, Shimada2021-wm}, but the optimization process can be time-consuming, unless certain compromises are assumed (\eg applying external non-physical forces \cite{Shimada2020-wv, Shimada2021-wm}). This disjoint process uses the kinematic body pose as an intermediate representation to communicate movement information to the humanoid controller. However, this communication layer can be fragile and adversely affected by the imitator's sensitivity to the intermediate representation. Thus, we propose to remove the kinematic pose layer and directly learn a mapping from the input to the control signals end-to-end.

\begin{figure*}
\vspace{-1.5mm}
\begin{center}
\includegraphics[width=1\textwidth]{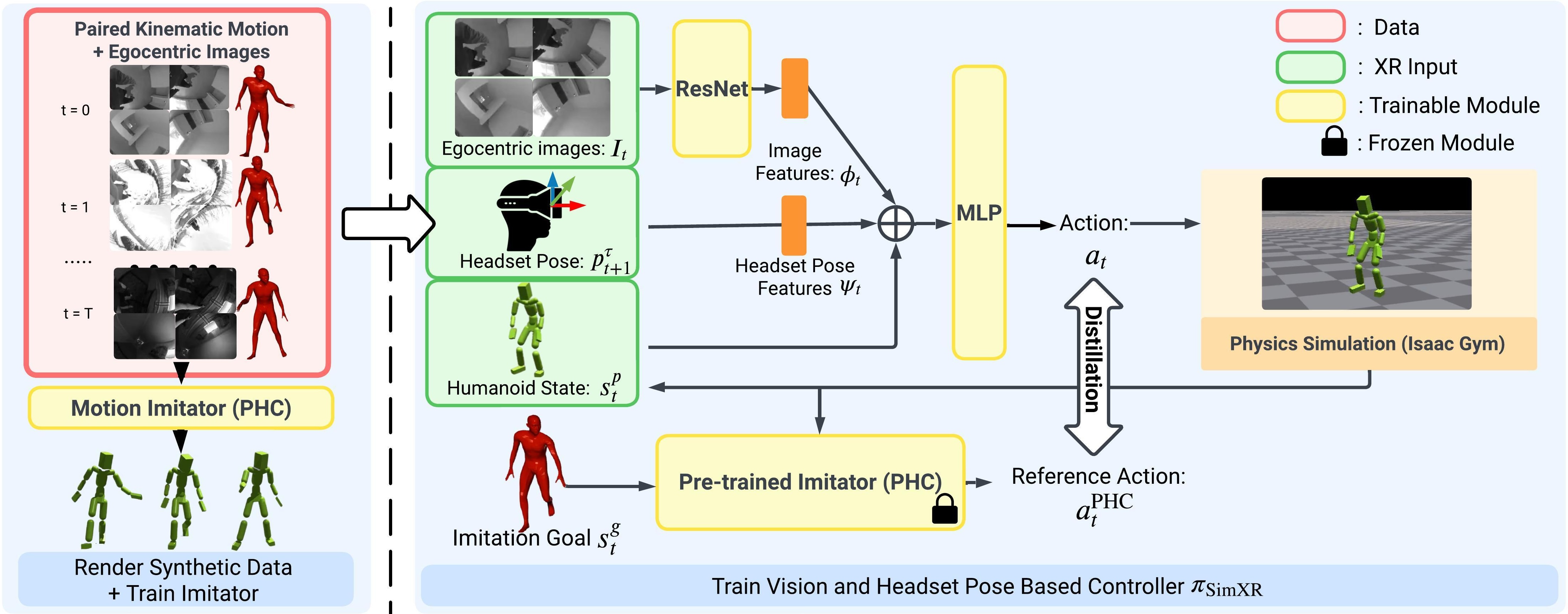}
\end{center}
\vspace{-5mm}
   \caption{\small{Our proposed $\name$ framework. From a large-scale human motion dataset, we first  train a motion imitator (PHC~\cite{Luo2023-ft}) and render synthetic images. Then, we train our vision and headset pose-based controller through distilling from the pretrained imitator.  }}
\vspace{-3.5mm}
\label{fig:archi}
\end{figure*}

\section{Approach}
At each time step, given the images $\imaget $ and the 6DoF pose $\traj$ captured by the headset, our task is to drive a simulated avatar to match the full body pose of the camera wearer $\simp$. We use monochrome SLAM cameras on XR headsets, producing images of dimension $\imaget \in \mathbb{R}^{V \times H \times W \times C}$ of $V$ views (2 views for Aria glasses, 4 for Quest) and $C = 1$ channels for monochrome images. Body pose $\simp \triangleq (\simr, \simt)$ consists of 3D joint rotation $\simr \in \reals^{J \times 6}$ (using the 6D rotation representation \cite{Zhou2019-lj}) and position $\simt \in \reals^{J \times 3}$ of all $J$ links on the articulated humanoid. Body velocities $\simvs$ are expressed as $\simv \triangleq (\simav, \simlv)$, consisting of angular $\simav \in \reals^{J \times 3}$ and linear velocities $\simlv \in \reals^{J \times 3}$. Throughout the paper, we use $\widehat{\cdot}$ to denote the ground-truth kinematic motion from Motion Capture (MoCap) and normal symbols without accents for values from the physics simulation. In Sec.\ref{sec:prelim}, we will first set up the preliminaries for humanoid control. Then, in Sec.~\ref{sec:syn}, we briefly describe our synthetic data generation pipeline. In Sec.~\ref{sec:simxr}, we describe our proposed method $\name$ and how to learn this controller.

\subsection{Preliminaries: Simulated Humanoid Control}
\label{sec:prelim}
Instead of directly regressing full-body pose, we use a simulated humanoid to ``act" inside physics simulation and read out its states as pose estimates. Following the general framework of goal-conditioned RL, we aim to obtain a goal-conditioned policy $\pi$ to control humanoids based on input from the headset. The framework is formulated as a Markov Decision Process (MDP) defined by the tuple ${\mathcal M}=\langle \mathcal{\bs S}, \mathcal{ \bs A}, \mathcal{ \bs T}, \rewardfunc, \gamma\rangle$ of states, actions, transition dynamics, reward function, and discount factor. Physics simulation determines the state $\state \in \mathcal{ \bs S}$ and the dynamics of the transition $\mathcal{ \bs T}$, where a policy computes the action $\action$. For humanoid control tasks, the state $\state$ contains the proprioception $\selfstate$ and the goal state $\goalstate$. Proprioception is defined as $\selfstate \triangleq (\simp, \simv)$, which contains the 3D body pose $\simp$ and the velocity $\simv$. The goal state $\goalstate$ is defined based on the task. Based on proprioception $\selfstate$ and the goal state $\goalstate$, the reward $r_t = \rewardfunc(\selfstate, \goalstate)$ is used to train the policy using RL (\eg PPO \cite{Schulman2017-ft}). If a reference action $\actiongt$ can be provided (often by a pre-trained expert), we can also optimize the policy $\pi$ through policy distillation~\cite{Rusu2015-tj, Ross2010-cc, Schmitt2018-ki}.

\paragraph{Physics-based Motion Imitation} Motion imitation is defined as the task of controlling a simulated humanoid to match a sequence of kinematic human motion $\refps$.
A motion imitator is formulated as follows: given the next-frame reference 3D pose $\refpn$, a policy $\policyphc(\action | \selfstate, \goalstateimitate)$ computes the joint actuation $\action$ to drive the humanoid to match $\refpn$. The goal state for motion imitation is defined as $\goalstateimitate \triangleq   (\refrn \ominus \simr, \reftn -  \simt, \bs{\hat{v}}_{t+1} -  \bs{v}_t, \bs{\hat{\omega}}_t -  \bs{\omega}_t, \refrn, \reftn)$, which contains the one frame difference between the reference and the current pose, normalized with respect to the heading of the current humanoid. 
A trained motion imitator could be used as a teacher and distill its motor skills for downstream tasks \cite{Merel2020-qm, Bohez2022-cm, Won2022-jy, Luo2023-er}, and we follow this paradigm and use an off-the-shelf RL-trained motion imitator (PHC \cite{Luo2023-ft}) as a teacher to learn the mapping between the XR headset sensor and the control signals.

\paragraph{Avatar Control using head-mounted Sensors} Following the above definition, the task of controlling a simulated humanoid to match egocentric observation from head-mounted sensors can be formulated as using a control policy $\policyours(\action | \selfstate, \imaget, \trajn)$ to compute the joint action $\action$ based on image $\imaget$, headset pose $\trajn$, and humanoid proprioception $\selfstate$ to match the headset wearer's body pose $\refps$.

\subsection{Synthetic Data for XR Avatar}
\label{sec:syn}
As paired motion data and XR cameras \& headset tracking is difficult to obtain, we use synthetic data to train our method when real data is not available. Specifically, since no data exist for Quest 2's SLAM camera configuration, we create a large-scale synthetic one. We render human motion from a large-scale internal MoCap dataset using the exact camera placement and intrinsic of the Quest 2 headset in the Unity game engine~\cite{Beta-ProgramUnknown-md}. The headset moves with the head movement of the wearer as if it is worn. Our motion dataset contains diverse poses ranging from daily activities to jogging, stretching, gesturing, sports, \etc, similar to AMASS~\cite{Mahmood2019-ki} but uses a different kinematic structure from SMPL \cite{Loper2015-ey}. During rendering, we randomize clothing, lighting, and background images (projected onto a sphere) for \textbf{every} frame for domain randomization. As a result, the methods trained using our synthetic data can be applied to real-world scenarios without additional image-based domain augmentation during training. We render RGB images and then convert them to monochrome. Fig. \ref{fig:archi} shows samples of our synthetic data. For more information on our synthetic data, please refer to the Supplement.

\subsection{Simulated Humanoid Control From Head-Mounted Sensors}
\label{sec:simxr}

\paragraph{Sensor Input Processing} At each frame, the head-mounted sensors provide the image $\imagetn$ and the 6DoF headset pose $\trajtn$ (here we use t+1 to indicate the incoming pose/image for tracking). The input image $\imagetn$ (which contains all monochrome views) is first processed with a lightweight image feature extractor $\resnet$ (\eg ResNet18 \cite{He2015-hk}) to compute image features: $\imagefeat = \resnet (\imagetn)$. All V camera views share the same feature extractor (Siamese network). We also replace all batch normalization layers \cite{Ioffe2015-oc} with group normalization \cite{Wu2018-ey} for training stability.

For the 6DoF headset pose $\traj$, we treat it as a virtual ``joint", where $\trajn \triangleq (\trajtn, \trajrn)$ contains the global rotation $\trajrn$ and translation $\trajtn$ of the headset. We use the humanoid head to track the pose of the headset by computing the rotation and translation difference between the head joint $\simp^{\text{Head}}$ and the headset pose $ \trajn$. This creates the headset pose feature: $ \trajfeat = (\simr^{\text{Head}} \ominus \trajrn, \simt^{\text{Head}} - \trajtn)$. The image feature and the headset pose features are then concatenated with proprioception $\selfstate$ to form the input to an MLP to compute joint actions, as illustrated in Fig.\ref{fig:archi}.

\paragraph{Online Distillation} Learning to control humanoids using RL requires a large number of simulation steps (\eg billions of environment interactions), and our early experiments show that directly training an image-based policy using RL is infeasible (see Sec.~\ref{sec:abla}). This is due to the large increase in input and model size (\eg four $120 \times 160 $ monochrome images versus $938$d imitation goal $\goalstateimitate$) that prohibitively slows down training steps. Therefore, we opt for online distillation and use a pre-trained motion imitator, PHC $\policyphc$, to teach our policy $\policyours$ via supervised learning. In short, we offload the sample-inefficient RL training to a low-dimensional state task (motion imitation) and use sample-efficient supervised learning for the high-dimensional image processing task. This is similar to the process proposed in PULSE \cite{Luo2023-er} with the important distinction that PULSE uses the same input and output for the student and teacher, while ours has drastically different input modalities (images and headset pose vs. 3D pose).

Concretely, we train our pose estimation policy $\policyours$ following the standard RL training framework: for each episode, given paired egocentric images $\images$, headset pose $\trajs$, and the corresponding reference full-body pose $\refps$, the humanoid is first initialized as $\refpzero$. Then, the policy $\policyours(\action | \selfstate, \imaget, \trajn)$ computes the joint actuation for the forward dynamics computed by physics simulation. By rolling out the policy in simulation, we obtain paired trajectories of $(\selfstates, \images, \trajs, \refps)$. Using the ground truth reference pose $\refps$ and simulated humanoid states $\selfstates$, we can compute the per-frame imitation target for our pretrained imitator $\goalstateimitate \leftarrow (\selfstate, \refpn)$. Then, using paired $(\selfstate, \goalstateimitate)$, we query PHC $\policyphc(\actionphc | \selfstate, \goalstateimitate)$ to calculate the reference action $\actionphc$. This is similar to DAgger \cite{Ross2010-cc}, where we use an expert to annotate reference actions for the student to learn from. To update $\policyours$, the loss is:
\begin{equation}
    \mathcal{L}  = \|\actionphc - \action \|^2_2, 
    \label{eq:loss}
\end{equation}
using standard supervised learning. In this way, our policy is trained end-to-end, where the image feature extractors $\resnet$ and the policy networks are directly updated using the gradient of $\mathcal{L}$. The learning process is also described in Alg.~\ref{alg:train}.

\begin{algorithm}[tb]
\footnotesize
  \caption{\small{Learn $\name$ via distillation}}\label{alg:train}
    
    \
    \textbf{Input:}  Ground truth paired XR sensor input and pose dataset $ \motiondata$, pretrained PHC $\policyphc$\,\;
         \While{not converged}{
            ${\bs M} \leftarrow \emptyset $ initialize sampling memory \,\;
                \While{${\bs M}$ not full}{
                   $\refps, \images, \selfstate, \leftarrow$ sample motion, images and initial state from $\motiondata$ \,\;
                    \For{$t \leftarrow 1...T$}{
                        $\bs a_t  \leftarrow \policyours(\action | \selfstate, \imaget, \trajn) $ \,\;
                        \tcp*[f]{compute humanoid action}\,\;
                         $\bs s_{t+1} \leftarrow \mathcal{T}(\bs s_{t+1} |\bs s_{t}, \bs a_t)$  \tcp*[f]{simulation}\,\;
                         $\goalstateimitate \leftarrow (\selfstate, \refpn)$ \,\;
                         \tcp*[f]{compute imitation target for PHC}\,\;
                         store $\selfstate, \goalstateimitate, \imaget, \trajn$ into memory ${\bs M}$ \,\;
                    }
                }
                $\actionphc \leftarrow \policyphc(\phcplusaction| \selfstate, \goalstateimitate)$ annotate collected states in ${\bs M}$ using $\policyphc$ \,\;
                
                $\policyours \leftarrow $ supervised update for $\policyours$ using pairs of $(\action, \actionphc)$ and Eq.\ref{eq:loss}. 
            }
   \vspace{-0.2cm}
\end{algorithm}

\section{Experiments}
\begin{table}[t]
\label{tab:dataset}
\centering
\resizebox{\linewidth}{!}{%
\begin{tabular}{rrr|rrr|rr}
\toprule
 \multicolumn{3}{c|}{Aria Glasses  }   & \multicolumn{5}{c}{Quest 2} 
\\
\midrule
  \multicolumn{3}{c|}{Aria Digital Twin \cite{Pan2023-ox}}    & \multicolumn{3}{c|}{Synthetic}  & \multicolumn{2}{c}{Real-world}
\\ 
\midrule
       Train & Test  & Annot.  & Train & Test & Annot. & Test &Annot. \\ \midrule
         {242 / 94k} & {64 /25k} & MoCap &  {4097 / 1758k} & {1072/458k} & MoCap & {10/40k} & Mono\\
\bottomrule 
\end{tabular}
}
\caption{\small{Dataset statistics and annotation source. MoCap: motion capture; Mono: monocular third-person pose estimation.}}
\vspace{-5mm}
\end{table}

\begin{figure*}
\vspace{-1.5mm}
\begin{center}
\includegraphics[width=1\textwidth]{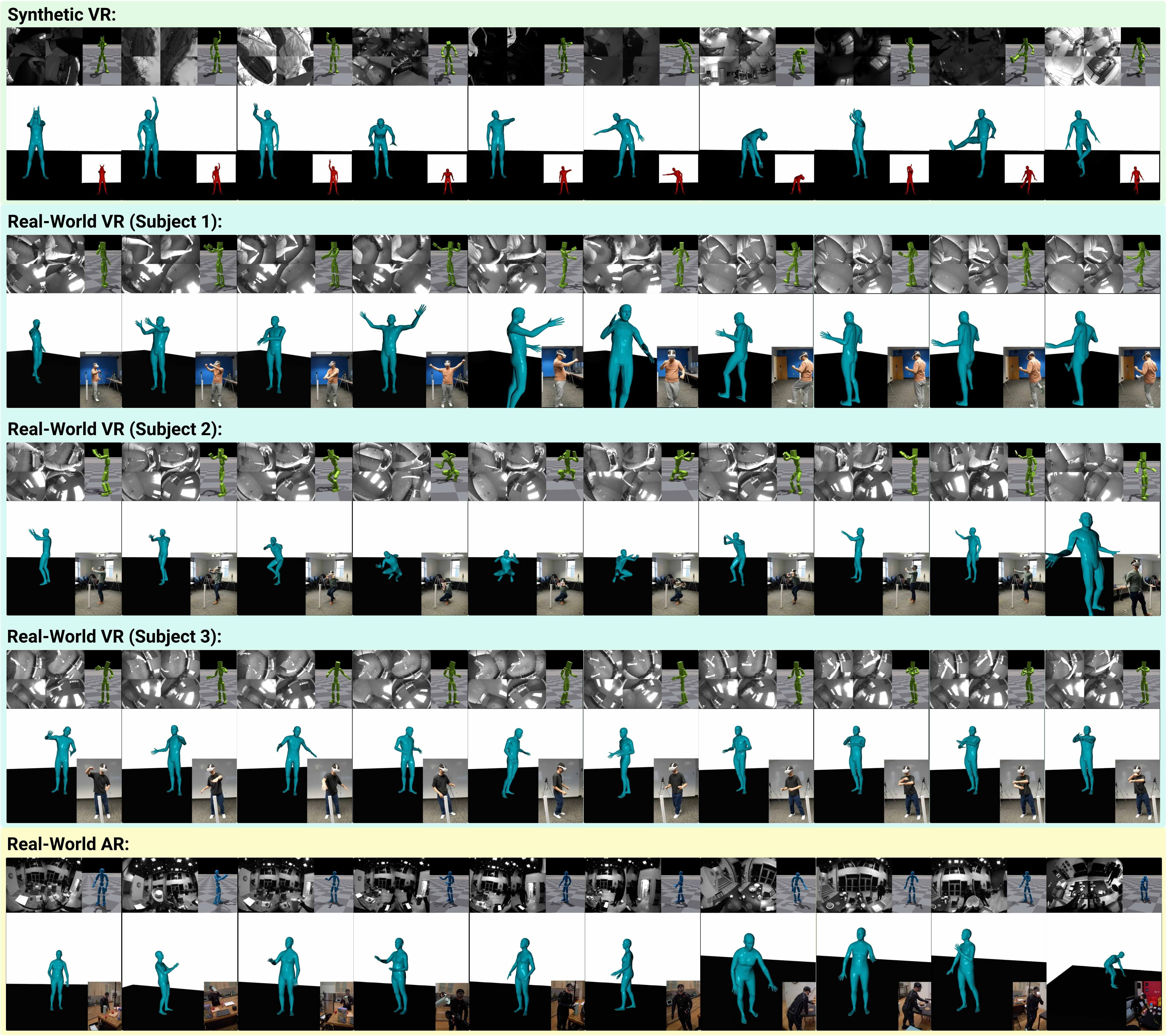}
\end{center}
\vspace{-5mm}
   \caption{\small{Qualitative results on synthetic and real-world AR/VR headset data. We visualize camera images, simulation, rendered mesh from simulation states, and third-person reference views. We show that our method can transfer to real-world data and handle diverse body poses including kicking, kneeling, \etc. For AR headset results, the third-person view is provided by another subject wearing a headset.  }}
\vspace{-3.5mm}
\label{fig:qual}
\end{figure*}
\begin{table*}[t]

\centering
\resizebox{\linewidth}{!}{%
\begin{tabular}{lrr|rrrrrr|rrrrr}
\toprule
\multicolumn{3}{c}{}  & \multicolumn{6}{c}{Synthetic-Test} & \multicolumn{4}{c}{Real-world} 
\\ 
\midrule
Method  & $\text{Size}$ & $\text{Physics}$ & $\text{Succ} \uparrow$ & $E_\text{g-mpjpe} \downarrow$ &  $E_\text{mpjpe} \downarrow $ & $E_\text{pa-mpjpe} \downarrow $ &  $\text{E}_{\text{acc}} \downarrow$  & $\text{E}_{\text{vel}} \downarrow$ & $\text{Succ} \uparrow$ & $E_\text{pa-mpjpe} \downarrow$ &    $\text{E}_{\text{acc}} \downarrow$  & $\text{E}_{\text{vel}} \downarrow$     \\ \midrule

\text{UnrealEgo \cite{Akada2022-zq}} & 554.5MB & \xmark &  {-} & {-} &  \textbf{56.9} &  {47.2}  & {54.5} & {32.5} & {-} &  {81.0} & {46.7} & {35.1} \\
\text{KinPoly-v \cite{Luo2021-gu}}&  98.9MB & \cmark &    {82.2\%} & {66.7}  & {63.5} & {42.8} & \textbf{4.4} & \textbf{5.8} & {9/10} & {83.0} & {7.0} & {11.2}\\

\text{Ours} & 59.7MB & \cmark &   \textbf{ 94.3\%} & \textbf{66.4} & {62.4}  & \textbf{40.0} & {6.5} & {8.3} & \textbf{10/10} & \textbf{73.0} &\textbf{6.8} & \textbf{10.5}\\

\bottomrule 
\end{tabular}}
\caption{\small{Pose estimation result on the test split ($458$k frames) of synthetic data and real-world captures ($40$k frames). Here, our MPJPE is computed as ``device-relative" instead of root-relative. }} 
\label{tab:quest}
\vspace{-5mm}
\end{table*}

\begin{table}[t]

\centering
\resizebox{\linewidth}{!}{%
\begin{tabular}{l|rrrrrr}
\toprule
\multicolumn{1}{c}{} & \multicolumn{6}{c}{ADT-Test} 
\\ 
\midrule
Method   & $\text{Succ} \uparrow$ & $\gmpjpe  \downarrow$ &  $\mpjpe \downarrow $ &  $\pampjpe$ &$\text{E}_{\text{acc}} \downarrow$  & $\text{E}_{\text{vel}} \downarrow$    \\ \midrule

\text{KinPoly \cite{Luo2021-gu}}  &  { 98.3\%} & {93.8} & {80.6}  & {60.8} & {5.9} & {9.5}\\
\text{UnrealEgo \cite{Akada2022-zq}}  &  {-} & {-} & {131.5} & {71.5}  & {26.7} & {20.4} \\
\text{Ours (headset-only)}  &  \textbf{100\%} & {120.7} & {120.6}  & {85.8} & {5.2} & {9.2} \\ %
\text{Ours}  &  \textbf{100\%} & \textbf{67.8} & \textbf{67.6}  & \textbf{47.7} & \textbf{4.6} & \textbf{7.3} \\

\bottomrule 
\end{tabular}}
\caption{\small{Pose estimation results on the ADT test set ($25$k frames).  }
} \label{tab:aria}
\vspace{-7.5mm}
\end{table}

\paragraph{Humanoids}
Due to the different annotation formats between our proposed synthetic dataset and public datasets, we use two different humanoids for our experiments, one for the Quest VR headset and one for Aria AR glasses. For Aria, we use a humanoid following the SMPL \citep{Loper2015-ey} kinematic structure with the mean shape. It has $24$ joints, of which $23$ are actuated, resulting in an actuation space of $\reals^{23 \times 3}$. Each degree of freedom is actuated by a proportional derivative (PD) controller, and the action $\action$ specifies the PD target. For the VR datasets, we use a humanoid that has $25$ joints (out of which $24$ are actuated). The imitator for SMPL-humanoid is trained on the AMASS~\cite{Mahmood2019-ki} dataset, while for the in-house humanoid it is trained on the same in-house motion capture used to create our synthetic dataset.

\paragraph{Datasets} To train our control policies, we require high-quality motion data paired with camera views, as training with low-quality data might lead to the simulated character picking up unwanted behavior. 
Thus, for pose estimation using the Quest headset, we train solely on synthetic data created using a large-scale in-house motion capture dataset. We randomly split the data using an 8:2 ratio, resulting in $1758$k frames for training and $458$k frames for testing. To test the performance of our method in real-world scenarios, we also collect a real-world dataset containing $40$k frames recorded by three different subjects. This dataset contains paired headset poses, SLAM camera images, and third-person images. We use the third-person images to create pseudo-ground truth using a SOTA monocular pose estimation method \cite{Sarandi2022-jh}. The real-world dataset contains daily activity motion common in VR scenarios, such as hand movements, boxing, kicking, \etc. For pose estimation using Aria glasses, we use the recently proposed Aria Digital Twin dataset (ADT) \cite{Pan2023-ox}. The ADT dataset contains indoor motion sequences captured using MoCap suits and 3D scene scanners. It contains $119$k frames that have paired skeleton and AR headset sensor output, recorded in a living room environment. This dataset only contains 3D keypoint annotations (no rotation), and we fit the SMPL body annotation to the 3D keypoints using a process similar to 3D-Simplify \cite{Bogo2016-kn, Pavlakos2019-fd}. We randomly split the ADT dataset for training ($94$k frames) and testing ($25$k frames). We do not use synthetic data for training the AR controller since there are available real-world ground-truth data.

\paragraph{Metrics} We report both pose- and physics-based metrics to evaluate our avatar's performance. We report the success rate ($\success$) as in UHC \cite{Luo2021-gu}, defined as: at \textit{every point} during imitation, the head joint is $< 0.5$m from the headset pose. For pose estimation, we report the \textbf{device-relative} (instead of root relative) per-joint position error (MPJPE) $\mpjpe$, global MPJPE $\gmpjpe$, and MPJPE after Procrustes analysis $\pampjpe$. Since $\pampjpe$ solves for the best matching scale, rotation, and translation, it is more suitable for our real-world dataset, where the scale and global position of the pseudo-ground-truth pose are noisy. To maintain the consistency of evaluation between the two humanoids, we select 11 common joints (head, left and right shoulders, elbows, wrists, knees, ankles) to report joint errors, rather than evaluating all joints as in the prior art~\cite{Luo2023-ft}. To test physical realism, we include acceleration $\acc$ (mm/frame$^2$) and velocity $\vel$ (mm / frame) errors. 

\paragraph{Baselines}  We adopt the SOTA vision-based pose estimation method UnrealEgo \cite{Akada2022-zq} as the main vision-based baseline. UnrealEgo uses a U-Net structure to first reconstruct 2D heatmaps from input images. Then, an autoencoder is used to lift the 2D heatmap to 3D keypoints. UnrealEgo predicts the 3D pose in the headset's coordinate system instead of the global one. To compare against a physics-based method, we reimplement KinPoly \cite{Luo2021-gu} and add image input to create KinPoly-v. We also remove its dependence on action labels and external forces (replacing UHC~\cite{Luo2021-gu} with PHC~\cite{Luo2023-ft}). KinPoly-v uses a two-stage method: first estimate full-body pose from images, and then feed them into a pretrained imitator to drive the simulated avatar. It uses a closed-loop system, where the simulated state is fed into the image-based pose estimator, making it ``dynamically regulated". KinPoly-v shares the same overall architecture as our method but uses kinematic pose as an intermediate representation to communicate with a pretrained imitator, instead of an end-to-end approach. 

\paragraph{Implementation Details} We use NVIDIA's Isaac Gym \cite{Makoviychuk2021-sx} for physics simulation. All monochrome images are resized to $120 \times 160$ for the Aria and Quest headsets during training and evaluation. All MLPs used are 6-layer with units [2048, 1536, 1024, 1024, 512, 512] and SiLU~\cite{Elfwing2017-is} activation. The networks for Quest and Aria share the same architecture, with the only difference being the first layer for the image feature extractor (processing 2 or 4 monochrome images). Due to the orders-of-magnitude increase in input size and network capacity (ResNet 18 vs. 6 layer MLPs), training image-based methods with simulation is around 10 times more resource intensive even using our lightweight networks. We train $\policyours$ for three days, collecting 0.1B environment interactions. For comparison, PHC trained for three days using RL collect around 2B environment interactions. After training, our estimation network and simulation run at $\sim$30 FPS.  We perform all the evaluation with a fixed body shape for both humanoids, as our pipeline does not infer or use any body shape information. For more implementation details, please refer to the supplement.

\subsection{Results}
As motion is best seen in videos, please refer to our supplement for extended visual results of our proposed method. 

\begin{table*}[h]

\centering
\resizebox{\linewidth}{!}{%
\begin{tabular}{r|rrrrrrrrrrrrr}
\toprule
\multicolumn{1}{c}{} & \multicolumn{11}{c}{Synthetic-Test, $\pampjpe \downarrow$ } 
\\ 
\midrule
 $\text{Method}$   & $\text{Head}$ & $\text{L\_Shoulder}$ & $\text{L\_Elbow}$ & $\text{L\_Wrist}$ & $\text{R\_Shoulder}$  & $\text{R\_Elbow}$ & $\text{R\_Wrist}$   & $\text{L\_Knee}$ & $\text{L\_Ankle}$ & $\text{R\_Knee}$ & $\text{R\_Ankle}$ \\ \midrule
UnrealEgo & \textbf{28.3} & {29.3} &  40.7 & 55.8& {29.2} & 38.2&  46.9& 53.0& 74.4 & 50.9& 72.8 \\
Ours & 30.2 & \textbf{25.1} & \textbf{31.8} & \textbf{46.1} & \textbf{24.5} & \textbf{31.4} &   \textbf{43.1} & \textbf{43.9}   &  \textbf{61.0} &  \textbf{43.8} &  \textbf{60.8}\\
\bottomrule 
\end{tabular}}
\vspace{-3mm}
\caption{\small{Per-joint error analysis on the $\pampjpe$} on the synthetic test set.} 
\label{tab:anla}
\vspace{-5mm}
\end{table*}

\paragraph{VR Headset Pose Estimation} In Table ~\ref{tab:quest} and Fig.~\ref{fig:qual}, we report the results of our synthetic and real-world dataset. Our method achieves comparable or better pose estimation results than both prior vision and vision + physics-based methods, estimating physically plausible full-body poses. Compared to UnrealEgo, we can see that $\name$ uses fewer parameters by an order of magnitude while achieving a better pose estimation result. While both UnrealEgo and $\name$ are single-frame methods (using no temporal architectures), $\name$ has significantly less jittering. This is due to the laws-of-physics as a strong prior, and the inherent temporal information provided by the simulation state. Note that UnrealEgo estimates pose in the device frame, so it provides a better headset-relative pose estimate in terms of $\mpjpe$. Our method directly estimates pose in the global coordinate system and does not benefit from aligning the head position (as can be seen in the small gap between global $\gmpjpe$ and local $\mpjpe$ joint errors. Compared to KinPoly-v, we can see that our method achieves better performance across the board. In KinPoly-v, the imitator, UHC \cite{Luo2021-gu}, uses an external non-physical force to help balance the humanoid and does not require any reference velocity $\refv$ as input. PHC does not use any non-physical forces and does require reference velocities as input, which creates additional difficulty in KinPoly-v. As a result, KinPoly-v does not perform well in both synthetic and real-world test sets, showing the limitations of the two-stage methods. Open-loop methods (where the policy does not have access to the kinematic state) may suffer less from the velocity estimation issue but introduce more disjoint between the kinematic and dynamic processes, as shown in KinPoly~\cite{Luo2021-gu}.  The relatively small gap between real-world and synthetic performance shows that our synthetic dataset is effective in providing a starting point for this challenging task.

\paragraph{AR Headset Pose Estimation}
Table~\ref{tab:aria} shows the test result on the ADT dataset. ADT is a much simpler dataset in terms of motion complexity, as it contains only walking, reaching, and daily activities. However, estimating pose from an AR headset is a harder task, as the viewing angle is much more challenging with the body not seen most of the time. Thus, UnrealEgo performs poorly on the dataset, unable to deal with unseen body parts. Our method can effectively leverage head movement and create plausible full-body motion. Due to similar issues in the VR headset case, KinPoly-v also does not perform well. To show that our method effectively uses the image input, Table \ref{tab:aria} also includes a headset-only variant of our approach, where we do not use any vision-based input. Comparing row 3 (R3) and R4, we can see that vision-based input indeed provides valuable information about the movement of the hands.

\subsection{Ablations and Analysis}
\label{sec:abla}
\begin{table}[t]

\centering
\resizebox{\linewidth}{!}{%
\begin{tabular}{cccc|rrrr}
\toprule
\multicolumn{2}{c}{} & \multicolumn{5}{c}{Synthetic-Test} 
\\ 
\midrule
 $\text{Vision}$ & $\text{Headset}$  & $\text{GN}$   & $\text{Distill}$& $\text{Succ} \uparrow$ & $\gmpjpe  \downarrow$ &    $\acc \downarrow$ & $\vel \downarrow$ \\ \midrule

\xmark & \cmark & \cmark & \cmark &   {73.0\%} & {118.9} & {8.8}  & {11.6}  \\ %
\cmark & \xmark & \cmark & \cmark &   {27.1\%} & {161.5}  & {6.8}  & \textbf{8.3}  \\ %
\cmark & \cmark & \xmark & \cmark &   {93.8\%} & {74.9} & {7.3} & {9.3}\\ %
\cmark & \cmark & \cmark & \xmark &   {35.8\%} & {226.1} & {9.6} & {9.9}  \\ %
\midrule
\cmark & \cmark & \cmark & \cmark &  \textbf{94.3\%} & \textbf{66.4} & \textbf{6.5} & \textbf{8.3}\\ %

\bottomrule 
\end{tabular}}
\caption{\small{Ablations on components of $\name$: without vision/headset input, not suing group norm, and no distillation. }} \label{tab:abla}
\end{table}

\begin{figure}
\vspace{-1.5mm}
\begin{center}
\includegraphics[width=0.45\textwidth]{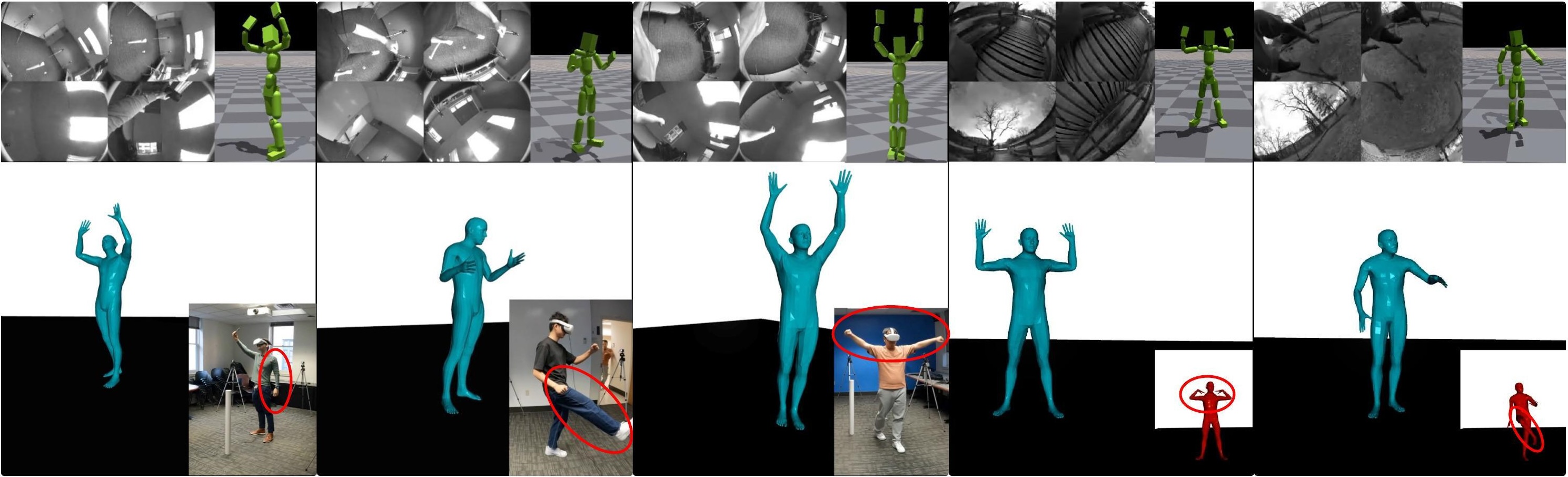}
\end{center}
\vspace{-5mm}
   \caption{\small{ Failure cases of our method: misplaced feet or hands.}}
\vspace{-3.5mm}
\label{fig:fail}
\end{figure}
In Table \ref{tab:abla}, we ablate the components of our method on the synthetic test set. Comparing R1, R2 and R5, we can see the importance of each of the two modalities: the vision signals provide most of the end-effector body movement signals, while the headset guides the body root motion. Without vision signals, the humanoid would achieve poor pose estimation results but can still achieve a reasonable success rate since the headset pose provides a decent amount of movement signals (R1). Without headset pose signals (R2), the humanoid will soon lose track of the head pose. In this case, the method needs to perform both SLAM and pose estimation from images, which requires special architectures. Comparing R3 and R5, we can see that the use of group normalization instead of batch normalization provides some boost in performance.  R4 shows that training from scratch using RL for vision-based methods without using distillation would require more efficient algorithms. 

To further analyze our pose estimation results, we report the per-joint values for $\pampjpe$ in Table \ref{tab:anla}. We can see that $\name$, similar to UnrealEgo, makes the most mistakes in end effectors such as the wrists and ankles. It performs worse in head alignment since it uses a humanoid to track the headset pose, but gains performance on lower-body joints such as ankles. This is due to the fact that $\name$ effective uses physics as a prior and can create plausible lower body movement based on input signals.

\section{Conclusion and Future Work}
\paragraph{Failure Cases}
In Fig.~\ref{fig:fail}, we visualize some failure cases of our method. As one of the first methods to control a simulated humanoid to match image observations from commercial headsets, it can misinterpret hand positions when they are not observed. Some kicking motion can also be ignored if the feet observation is blurry (due to clothing color, lighting \etc).  We also notice that, in some cases, the simulated humanoid movement might lag behind the real-world images, especially when hands come in and out of view. This can be attributed to the humanoid being conservative and not committing to movement until the body part is fully visible. In the videos, we can observe the humanoid stumbling and dragging its feet to remain balanced, especially when the headset moves too quickly. 

\paragraph{Conclusions}
We propose $\name$ to control simulated avatars to match sensor input from commercially available XR headsets. We propose a simple, yet effective, end-to-end learning framework to learn to map from headset pose and camera input from XR headsets to humanoid control signals via distillation.  To train our method and facilitate future research, we also propose a large-scale synthetic dataset for training and a real-world dataset for testing, all captured using standardized off-the-shelf hardware. Training only using synthetic data, our lightweight networks can control simulated avatars in real-world data capture with high accuracy in real-time. Future directions include adding auxiliary loss during training to improve the accuracy of pose estimation, incorporating temporal information, and using scene-level information for better pose estimates. 

\newpage
\paragraph{Acknowledgement} We thank Zihui Lin for her help in making the plots in this paper. Zhengyi Luo is supported by the Meta AI Mentorship (AIM) program.
 
{
    \small
    \bibliographystyle{ieeenat_fullname}
    \bibliography{main}
}

\clearpage
\appendix{   
    \hypersetup{linkcolor=black}
    \begin{Large}
        \textbf{Appendix}
    \end{Large}
    \etocdepthtag.toc{mtappendix}
    \etocsettagdepth{mtchapter}{none}
    \etocsettagdepth{mtappendix}{subsection}
    \newlength\tocrulewidth
    \setlength{\tocrulewidth}{1.5pt}
    \parindent=0em
    \etocsettocstyle{\vskip0.5\baselineskip}{}
    \tableofcontents
}

\renewcommand\thefigure{A.\arabic{figure}}  
\setcounter{figure}{0}    
\renewcommand\thetable{A.\arabic{table}}  
\setcounter{table}{0}    

\section*{Appendices}
\section{Introduction}
In this supplement, we provide additional details about $\name$ that are left out of the main paper due to space constraints. Specifically, in Sec.~\ref{sec:supp_site}, we describe the contents of our supplementary site and videos. In Sec.~\ref{sec:supp_impl}, we discuss the implementation details of our proposed synthetic dataset (Sec~\ref{sec:supp_impl_syn}), our proposed method $\name$ (Sec~\ref{sec:supp_impl_simxr}), pretrained imitators that act as teachers (Sec~\ref{sec:supp_impl_imitator}),  KinPoly-v (Sec~\ref{sec:supp_impl_kinpolyv}), and UnrealEgo (Sec~\ref{sec:supp_impl_unrealego}). Finally, in Sec.~\ref{sec:supp_abla}, we provide some additional ablations (such as using recurrent networks) and analysis of failure cases. Since motion is best seen in videos, we strongly encourage our readers to view the provided videos for a qualitative analysis of our method. \textbf{All data and models will be released. }

\section{Supplementary Site \& Videos}
\label{sec:supp_site}
In the supplement site, we provide an extensive qualitative evaluation of our method. We visualized all three subjects and \textbf{full} sequences of our real-world Quest 2 data capture, as well as results from the synthetic dataset. From the videos, we can see that our end-to-end method can follow the headset wearer's body motion closely in a physically plausible fashion. We also show results on AR/Aria glasses and compare with SOTA vision-based and physics-based methods. Last but not least, we visualize failure cases of our method.

\section{Implementation Details}
\label{sec:supp_impl}
\subsection{Synthetic Data Generation}
\label{sec:supp_impl_syn}
\begin{figure}
\vspace{-1.5mm}
\begin{center}
\includegraphics[width=0.5\textwidth]{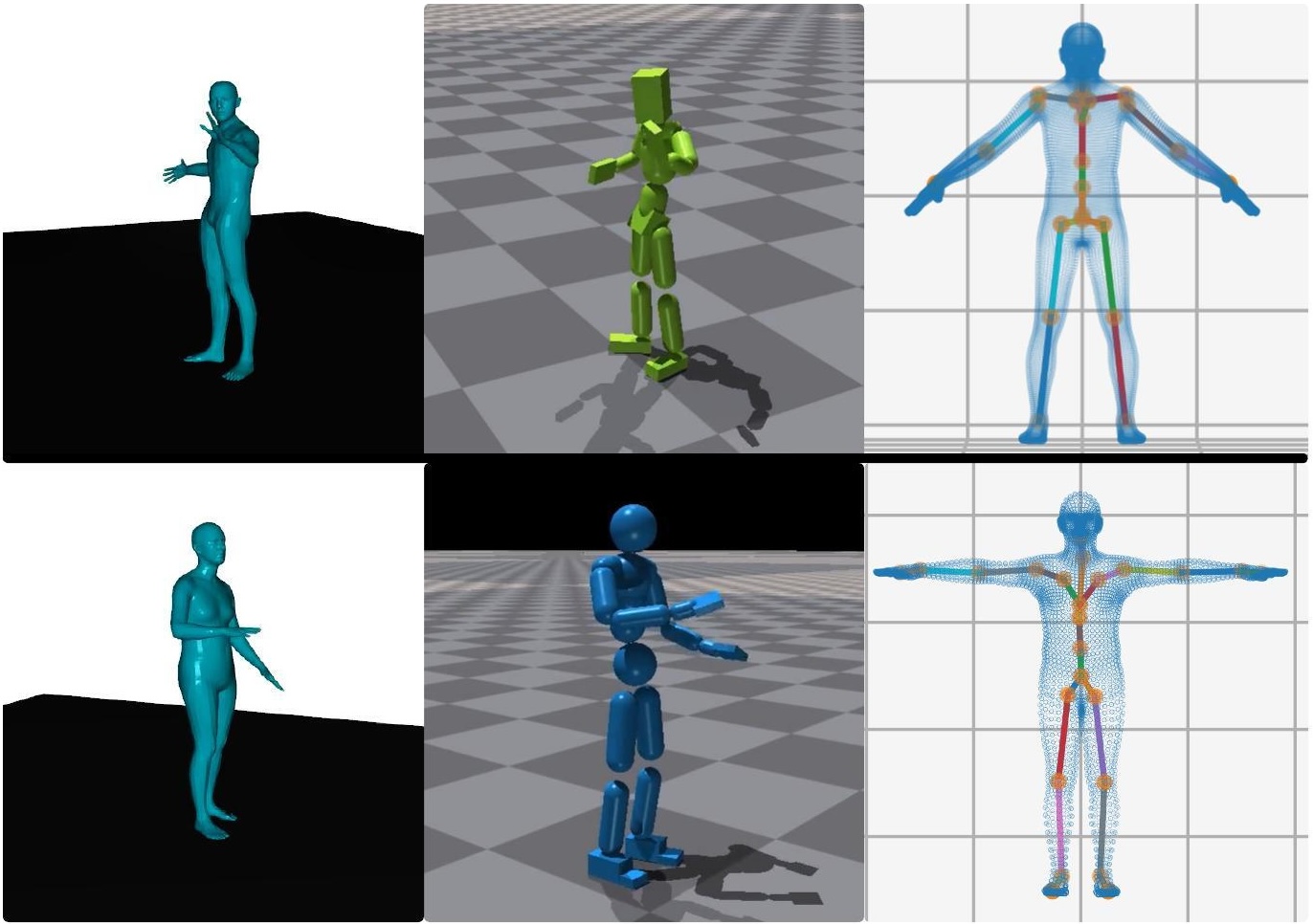}
\end{center}
\vspace{-5mm}
   \caption{\small{The two human models we use, their rendered mesh, simulated humanoid, and kinemaitc structure.  \textit{(Top)}: our in-house humanoid with 24 DOF. \textit{(Bottom)}: SMPL humanoid with 23 DOF. 
   }}
\vspace{-3.5mm}
\label{fig:syn_humanoid}
\end{figure}
\paragraph{Humanoid Kinematic Structure}
To create the synthetic egocentric data, we use an internal human mesh model similar to SMPL~\cite{Loper2015-ey}. Given kinematic body rotations and scale parameters, we can create its corresponding mesh as shown in Fig.~\ref{fig:syn_humanoid}. We use a process similar to that of the SMPL humanoid to create an IsaacGym compatible humanoid for the in-house human mesh model.

\paragraph{MoCap Dataset}
To generate synthetic data, we use a large-scale internal MoCap dataset consisting of 130 subjects and $> 1300$ capture sessions. The motion capture dataset contains a large number of daily activities (walking, running, gesturing, yoga, dancing, balancing, sitting, interaction with objects \etc). We remove sequences that contain human-object interactions that are not possible to mimic without simulating the objects (such as sitting on chairs). Since each capture session contains a long sequence of motion, we further divide the sessions into sequences that contain around $\sim$450 frames of motion, resulting in a total of $5169$ sequences for training and testing.

\begin{figure*}
\vspace{-1.5mm}
\begin{center}
\includegraphics[width=1\textwidth]{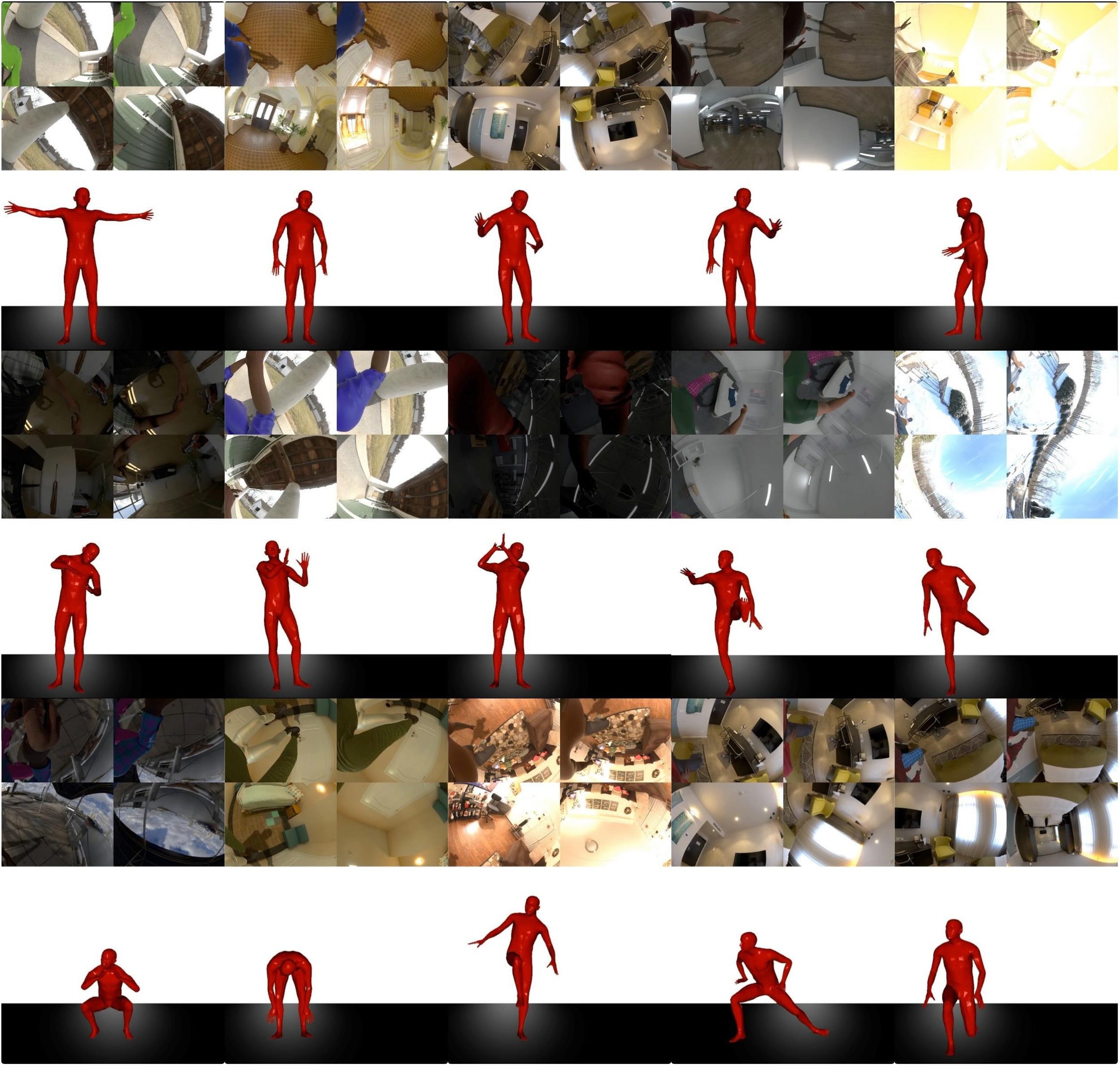}
\end{center}
\vspace{-5mm}
   \caption{\small{Sample synthetic data with various poses. Here we include the original rendered RGB images for demonstration purposes. We randomize the actor's clothes, background, lighting at every frame.  
   }}
\vspace{-3.5mm}
\label{fig:syn_data}
\end{figure*}

\paragraph{Rendering Pipeline} As mentioned in the main paper, rendering is done using the exact placement, intrinsic, and distortion of the cameras as the Quest 2 headset. The Unity~\cite{Beta-ProgramUnknown-md} game engine is used for rendering. In Fig.~\ref{fig:syn_data}, we show examples of raw RGB images rendered using our pipeline. In each frame, we randomize the clothing, lighting, and background of the subjects as domain randomization. The background is rendered using a random image projected onto a skybox. We render each frame of motion from the MoCap dataset in 30 FPS. Each RGB image is rendered in a $640 \times 480 \times 3$ resolution, and we convert the images to monochrome and shrink them to $160 \times 120 \times 1$ for training and testing $\name$.

\begin{table}[h!]
\vspace{-2mm}
\centering
\resizebox{\linewidth}{!}{%
\begin{tabular}{r|rrrrrrrr}
\toprule
\multicolumn{1}{c}{} & \multicolumn{6}{c}{Synthetic-Test, $\pampjpe \downarrow$} 
\\ 
\midrule
 $\text{}$   & $\text{Head}$ & $\text{L\_Shoulder}$ & $\text{L\_Elbow}$ & $\text{L\_Wrist}$ & $\text{R\_Shoulder}$  & $\text{R\_Elbow}$ \\ \cmidrule{2-7}
In-frame \% & 0.0\% & 4.0\% & 61.3 \% & 92.4 \% &18.1 \% & 75.8\% \\ 
 In-frame / Out-frame  & - / 30.2 & 28.0 / \textbf{25.0} & \textbf{30.7} / 33.2& \textbf{42.7} / 86.3 & 25.6 / \textbf{24.3} & \textbf{30.4} / 34.3 \\ \midrule
  & $\text{R\_Wrist}$   & $\text{L\_Knee}$ & $\text{L\_Ankle}$ & $\text{R\_Knee}$ & $\text{R\_Ankle}$ \\ \cmidrule{2-7}
  In-frame \% &  96.5 \%& 99.0 \% & 98.4 \% & 99.4\% & 98.9\% \\
   In-frame / Out-frame  & \textbf{41.1} / 94.2  & 43.9 / \textbf{39.9}   &  \textbf{60.4} / 74.4 & \textbf{ 43.7 } / 45.5&  \textbf{60.3} / 72.7\\

\bottomrule 
\end{tabular}}
\vspace{-2mm}
\caption{\small{Error analysis based on joint in-frame status. ``In-frame" is not equivalent to ``visible" due to self-occlusion.  }}
\label{tab:resp_analysis}
\vspace{-5mm}
\end{table}

 \begin{figure}

\vspace{-1.5mm}
\begin{center}
\includegraphics[width=0.45\textwidth]{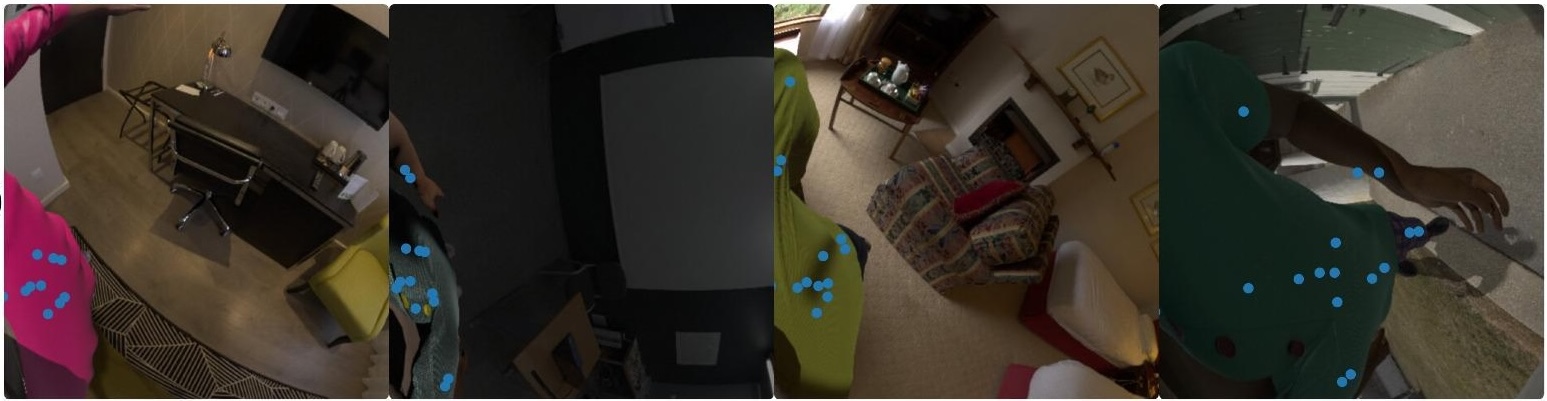}
\end{center}   
\vspace{-6mm}
   \caption{\small{ Self-occlusion visualization; blue dots are keypoints. }}
\label{fig:resp}
\vspace{-7mm}
\end{figure}
\paragraph{Visibility Analysis}
Here we conduct a visibility analysis on our generated synthetic data. 
As accounting for self-occlusion involves reprocessing the synthetic dataset and conducting ray-marching for each joint on the clothed mesh to ascertain visibility, we opt to use the ``in-frame" (whether the joint is in one of the camera frames) statistics to approximate visibility. This measure is a reliable visibility indicator for upper body joints, but less so for the lower body, where self-occlusion and extreme viewpoints are more prevalent (see Figure~\ref{fig:resp}). From Table~\ref{tab:resp_analysis} we can see that the shoulder and elbow joints are frequently outside the frame, while the wrist and leg joints often remain within the frame. For the shoulder, elbow, and knee joints, their in-frame and out-frame results are similar, since they are closer to the torso. For the body extremities (wrists and ankles), there is a clear gap, as the largest errors occur out of the frame.

\subsection{Details about $\textbf{\name}$}
\label{sec:supp_impl_simxr}

\paragraph{Body Shape Used for Evaluation} We conduct all our training and evaluation using a fixed body shape for both of our humanoids (SMPL and in-house). In other words, we use the mean body shape for SMPL and a fixed body shape for our internal humanoid and do not vary bone lengths between different motion sequences or subjects. Since our framework does not involve any intermediate representations such as 3D keypoints or poses, $\name$ is scale invariant.  When conducting real-world evaluations, we simply adjust the height of the headset pose to match the standing head positions of the mean body shape. This is done in a calibration phase in which the subject is standing still. This process is effective as $\name$ can estimate the pose for the three subjects who have different heights. Notice that our imitator can be trained to handle different body shapes, but we opt out of this option as estimating body shape from the distorted egocentric views is still an unsolved problem. 

\paragraph{Training Process} The training process for $\name$ is similar to training a motion imitator, with the distinction being that we provide images and headset pose as input instead of full-body reference pose. To better learn harder motion sequences, we use the same hard-negative mining process proposed in PHC \cite{Luo2023-ft} and PULSE \cite{Luo2023-er}. Concretely, during training, given the full motion and image dataset $\motiondata$, we evaluate the current policy on the full dataset and pick the sequences that the policy fails to form $\hardmotiondata$. We keep updating $\hardmotiondata$ at intervals until the success rate no longer increases. The hyperparameters for training $\name$ can be found in Table \ref{tab:supp_hyper}. 

\begin{table}[t]
\centering
\resizebox{\linewidth}{!}{%
\begin{tabular}{lcccccc}
\toprule
   & Batch Size & Learning Rate   & \text{\# of samples} &  $\text{image size}$  & $\text{Image-latent}$ 
  \\ \midrule
$\name$     & 1024 & $5 \times 10^{-4}$ &  $\sim 10^{8}$   &  $160 \times 120$  & 512 
\\ \midrule
& Batch Size & Learning Rate   & \text{\# of samples}
  \\ \midrule
PHC   & 3072    &  $2 \times 10^{-5}$   & $\sim 10^{10}$   & \\
\bottomrule 
\end{tabular}}\\ 
\caption{Hyperparameters for $\name$ and PHC. Due to the increase in input size, $\name$ is trained with significantly less samples than PHC and requires distillation. } 
\label{tab:supp_hyper}
\end{table}

\subsection{Details about Pretrained Imitators}
\label{sec:supp_impl_imitator}
For the SMPL humanoid (AR / Aria glass experiments), we use an off-the-shelf motion imitator, PHC \cite{Luo2023-ft}, trained on the AMASS dataset. We do not make any additional modifications to PHC, since the motion in the ADT dataset is relatively simple. For the in-house humanoid and VR / Quest experiments, we train an imitator using the same training procedure and hyperparameters provided in the PHC implementation. We train the imitator using the training sequences from the internal MoCap dataset, achieving the imitation performance shown in Table \ref{tab:quest_im}. We can see that the imitator has a high success rate and a low joint error on the training data, which means that it is suitable to be used as a teacher for downstream tasks.

\begin{table}[t]

\centering
\resizebox{\linewidth}{!}{%
\begin{tabular}{l|rrrrrr}
\toprule
\multicolumn{1}{c}{}  & \multicolumn{6}{c}{Synthetic-Train} 
\\ 
\midrule
Method  & $\text{Succ} \uparrow$ & $E_\text{g-mpjpe} \downarrow$ &  $E_\text{mpjpe} \downarrow $ & $E_\text{pa-mpjpe} \downarrow $ &  $\text{E}_{\text{acc}} \downarrow$  & $\text{E}_{\text{vel}} \downarrow$    \\ \midrule
\text{PHC} &    { 99.8\%} & 25.6 & {20.4}  & {15.5} & {2.4} & {3.5} \\
\bottomrule 
\end{tabular}}
\caption{\small{Motion imitation result by the pretrained imitator on the in-house MoCap dataset. }} 
\label{tab:quest_im}
\end{table}

\subsection{Details about KinPoly-v}
\label{sec:supp_impl_kinpolyv}

We adapt KinPoly \cite{Luo2021-gu} to also consume images as input for egocentric pose estimation. In KinPoly, a policy is learned to produce kinematic full-body poses $\kinpn$ based on the pose of the headset, which is then fed to an external force based motion imitator (UHC) for physics-based imitation. Comparing KinPoly to previous methods that use both an imitator and a pose estimator (\eg SimPOE \cite{Yuan2021-rl}), the main difference is whether the pose estimator is aware of the simulated humanoid state. In prior art, the pose estimator is not conditioned on the simulation state and operates independently from the physics simulation. This creates an open-loop system where the pose estimator can estimate a pose that drifts far away from the simulation state, leading the imitator to fall. KinPoly aims to create a closed-loop system where the pose estimator also takes the simulation state into consideration. This methodology is also used in EmbodiedPose \cite{Luo2022-ux}. However, the main problem with KinPoly is that, while it is relatively easy to output the correct reference kinematic pose $\kinpn$ for the current timestep, it is difficult to compute the correct velocities $\Tilde{\dot{\bs{q}}}_{t + 1}$. Due to the exceptional capabilities of the non-physical forces, UHC does not require reference velocities as input. Concretely, UHC's goal state is defined as $\goalstateimitate \triangleq   (\refrn \ominus \simr, \reftn -  \simt)$, which does not contain any velocity information. As a result, KinPoly's kinematic policy only needs to predict body pose, but not velocity, which simplifies the learning problem. However, since PHC does not use any external forces \cite{Yuan2020-fp}, it requires accurate reference velocities as input. 

To remove the dependency on external forces (change from UHC to PHC), we experimented with two forms of velocity prediction. The first approach is to compute the velocity as a finite difference between consecutive frames of the predicted reference poses: $\Tilde{\dot{\bs{q}}}_{t + 1} = \kinpn - \kinp$. This formulation is problematic as large jumps in predicted poses can result in large velocities, which in turn lead to the imitator falling. Another approach is to directly predict the velocity as an output, which turns out to be more stable. We use this version as our implementation for KinPoly-v. However, this approach still suffers from inaccurate velocity prediction, as can be seen in the supplement videos: when the motion becomes faster and more dynamic (such as sports movement or jogging), it becomes difficult to predict the correct velocities for the motion imitator. This can also be a result of the image input, which can be noisy and detrimental to the network learning a good velocity prediction. 

\begin{figure}
\begin{center}
\includegraphics[width=0.5\textwidth]{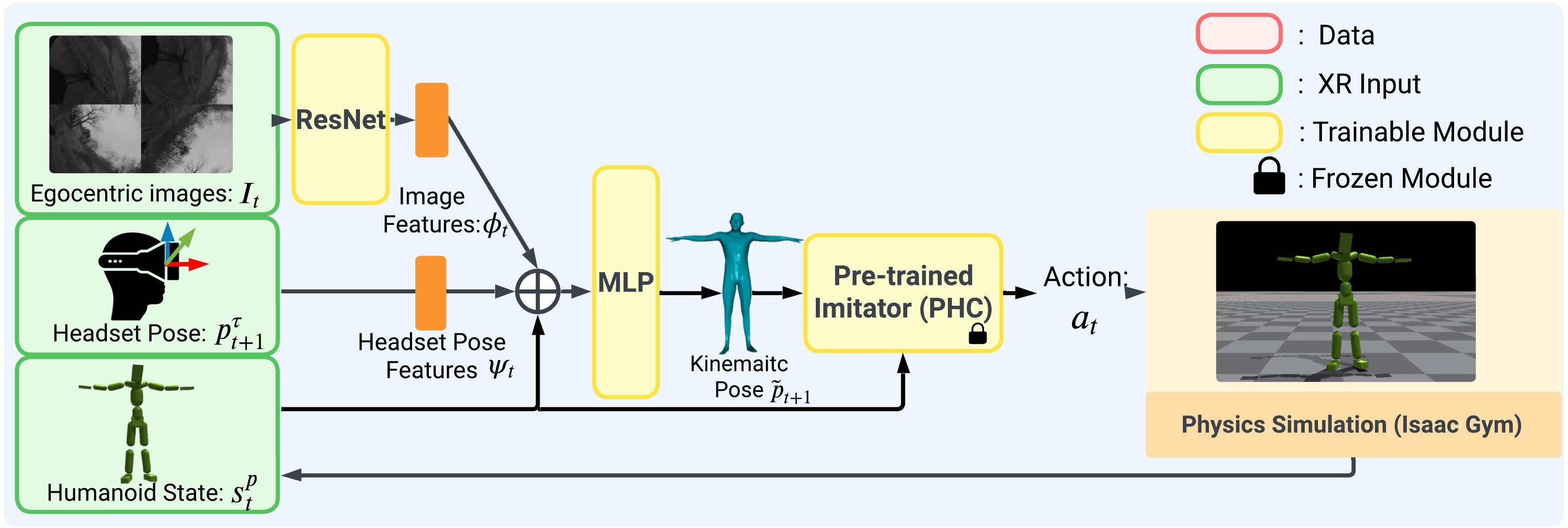}
\end{center}
\vspace{-5mm}
   \caption{\small{ KinPoly-v's network architecture. Different from distilling from a pretrained imitator, KinPoly-v outputs kinematic pose to the pretrained imitator for physcis-based motion imtiation. 
   }}
\label{fig:kinpoly}
\end{figure}

\paragraph{Network Architecture} KinPoly-v shares the same input and network architecture as $\name$, with the only distinction being the output: KinPoly-v outputs the kinematic pose $\kinpn$ rather than the PD targets $\action$ for the joints. KinPoly-v's architecture can be found in Fig.\ref{fig:kinpoly}

\subsection{Details about UnrealEgo}
\label{sec:supp_impl_unrealego}
We use the official UnrealEgo implementation, and pick the ResNet18 version with imagenet initialization for a fair comparison with $\name$. To be compatible with monochrome images, we replace the CNN layer with a single-channel convolutional layer and keep the Siamese network structure. We follow the official implementation and first train the 2D heatmap estimation network. Then, using the frozen heatmap estimation network, we train a 3D pose estimator based on the 2D heatmap input. We train the networks for three days to convergence, using a similar compute budget as training $\name$.

\section{Additional Ablations and Failure Cases}
\label{sec:supp_abla}
\begin{table}[t]

\centering
\resizebox{\linewidth}{!}{%
\begin{tabular}{l|rrrrrr}
\toprule
\multicolumn{1}{c}{} & \multicolumn{6}{c}{Synthetic-Test} 
\\ 
\midrule
Method   & $\text{Succ} \uparrow$ & $\gmpjpe  \downarrow$ &  $\mpjpe \downarrow $ &  $\pampjpe$ &$\text{E}_{\text{acc}} \downarrow$  & $\text{E}_{\text{vel}} \downarrow$    \\ \midrule

\text{Ours (GRU)}  &  {91.6\%} & {78.2} & {73.7}  & {52.8} & {8.8} & {10.4} \\ %
\text{Ours}  &   \textbf{ 96.2\%} & \textbf{69.0} & \textbf{65.2}  & \textbf{43.2} & \textbf{6.8} & \textbf{8.7} \\

\bottomrule 
\end{tabular}}
\caption{\small{Additional ablation on using recurrent architecture }
} \label{tab:rnn}
\end{table}

\paragraph{Recurrent Networks} Currently, $\name$ is a per-frame model without using any temporal model architecture. $\name$ does rely on temporal information in the form of a simulation state and estimates temporally coherent motion by jointly considering humanoid state and input images. Based on the intuition that incorporating recurrent networks is essential to help robots complete tasks \cite{Zhuang2023-nf}, we also tried recurrent architecture in our early experiments. We tested a simple GRU-based \cite{Cho2014LearningPR} architecture with 512 hidden units, and forms a lightweight Conv-LSTM~\cite{Sainath2015ConvolutionalLS}. Table \ref{tab:rnn} shows that the use of a recurrent network does not offer an immediate performance increase. We hypothesize that, for a pose estimation task with dense per-frame input, a recurrent network may not be necessary to ensure good performance. Further investigation is needed to better leverage the temporal coherence in videos.

\paragraph{Additional Failure Cases} In our supplementary videos, we visualize the common failure cases of $\name$. Being one of the first methods to drive simulated avatars from images and headset pose input from XR headsets, $\name$ shows the feasibility of training such a network, but it is still far from perfect. 
\begin{itemize}
    \item We can observe that the humanoid stumbles and drags its feet to stay balanced when moving around, which is caused by ambiguity in movement signals. Since our humanoid has no information about the future movements of the camera wearer, it adopts the foot-dragging behavior to be cautious and stay balanced. 
    \item Accurate foot movement can still be challenging: while $\name$ can estimate kicking and raising feet, it can also miss raised feet due to the challenging viewing angle. The foot can be barely visible even when raised, and the color of the garment can create additional ambiguities. 
    \item We can observe that when the hands are held perfectly still, the humanoid can have micro movements due to inaccuracy in inferring the body poses. Tackling this issue is challenging, as the movement of the headset can cause the hands to move in the camera space but not in the global space, and differentiating between the two requires further investigation.
    \item  The humanoid can also have erroneous hands movement from time to time, erecting the hand quickly and putting them down due to image noise and occlusion.
    \item Another source of inaccuracy is fast and sporty movement, where the humanoid can lag behind in performing the actions or fall down. 
\end{itemize}

  In the future, our aim is to incorporate a larger MoCap and synthetic datasets to improve the robustness of the controller. Introducing auxiliary pose estimation losses during training could also improve $\name$.

\end{document}